\documentclass[10pt,twocolumn,letterpaper]{article}

\usepackage[pagenumbers]{wacv} 

\usepackage{graphicx}
\usepackage{amsmath}
\usepackage{amssymb}
\usepackage{booktabs}
\usepackage{lipsum}
\usepackage{multirow}
\usepackage{ulem}
\usepackage{enumitem}
\usepackage{soul}

\usepackage{pifont}
\newcommand{\cmark}{\ding{51}}%
\newcommand{\xmark}{\ding{55}}%
\usepackage[pagebackref,breaklinks,colorlinks]{hyperref}

\usepackage[accsupp]{axessibility}
\usepackage[capitalize]{cleveref}
\Crefname{section}{Section}{Sections}
\Crefname{table}{Table}{Tables}
\Crefname{figure}{Figure}{Figures}
\usepackage{array}
\usepackage{xcolor}
\usepackage{soul}
\def\wacvPaperID{1060} 
\def\confName{WACV}
\def\confYear{2024}

\begin{document}

\title{I-AI: A Controllable \& Interpretable AI System \\for Decoding Radiologists' Intense Focus for Accurate CXR Diagnoses}

\author{Trong Thang Pham\\
AICV Lab, University of Arkansas\\
Fayetteville, AR, USA 72703\\
{\tt\small tp030@uark.edu}
\and
Jacob Brecheisen\\
AICV Lab, University of Arkansas\\
Fayetteville, AR, USA 72703\\
{\tt\small jmbreche@uark.edu}
\and
Anh Nguyen\\
University of Liverpool\\
Liverpool, UK\\
{\tt\small anh.nguyen@liverpool.ac.uk}
\and
Hien Nguyen\\
University of Houston\\
Houston, Texas, USA 77204\\
{\tt\small hvnguy35@central.uh.edu}
\and
Ngan Le\\
AICV Lab, University of Arkansas\\
Fayetteville, AR, USA 72703\\
{\tt\small thile@uark.edu}
}
\maketitle

\begin{abstract}
\vspace{-1em}

In the field of chest X-ray (CXR) diagnosis, existing works often focus solely on determining where a radiologist looks, typically through tasks such as detection, segmentation, or classification. However, these approaches are often designed as black-box models, lacking interpretability. In this paper, we introduce Interpretable Artificial Intelligence (I-AI) a novel and unified controllable interpretable pipeline for decoding the intense focus of radiologists in CXR diagnosis. Our I-AI addresses three key questions: where a radiologist looks, how long they focus on specific areas, and what findings they diagnose. By capturing the intensity of the radiologist's gaze, we provide a unified solution that offers insights into the cognitive process underlying radiological interpretation. Unlike current methods that rely on black-box machine learning models, which can be prone to extracting erroneous information from the entire input image during the diagnosis process, we tackle this issue by effectively masking out irrelevant information. Our proposed I-AI leverages a vision-language model, allowing for precise control over the interpretation process while ensuring the exclusion of irrelevant features.


To train our I-AI model, we utilize an eye gaze dataset to extract anatomical gaze information and generate ground truth heatmaps. Through extensive experimentation, we demonstrate the efficacy of our method. We showcase that the attention heatmaps, designed to mimic radiologists' focus, encode sufficient and relevant information, enabling accurate classification tasks using only a portion of CXR. The code, checkpoints, and data are at \url{https://github.com/UARK-AICV/IAI}.

\vspace{-2em}


\end{abstract}



\normalem

\section{Introduction}
\label{sec:intro}



\begin{figure*}[t]
    \centering
    \includegraphics[width=\linewidth]{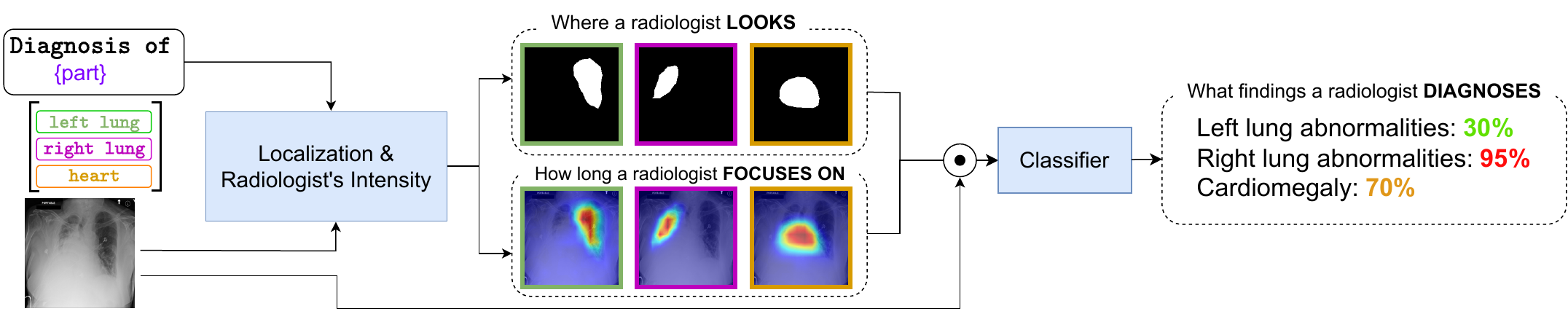}
    \vspace{-0.2in}
    \caption{The overall pipeline of our proposed I-AI framework to detect radiologist's intense focus for accurate CXR diagnoses. The model takes an CXR and anatomical prompt of a particular area of diagnosis as its inputs and outputs the answers of three key questions: where a radiologist looks, how long a radiologist focuses on specific areas, and what findings a radiologist diagnoses. $\odot$ is the Hadamard product.}
    \label{fig:overview}
\end{figure*}
\begin{table*}[t]
\centering
\caption{Model capacity comparison between our proposed method and related approaches.}
\resizebox{.7\textwidth}{!}{
\begin{tabular}{lccccc}
\toprule
\textbf{Methods} & \textbf{Localization}  & \textbf{Diagnosis} & \textbf{Intensity} & \textbf{Interpretability} & \textbf{Controllability} \\
\midrule
ChexNet~\cite{rajpurkar2017chexnet}  & \xmark & \cmark  & \xmark & \xmark &  \xmark \\
van Sonsbeek\cite{van2023probabilistic} & \xmark & \cmark  & \xmark & \xmark &  \xmark \\
Grad-CAM~\cite{selvaraju2017grad}  & \cmark & \xmark  & \cmark & \xmark &  \xmark \\
Grad-CAM++~\cite{chattopadhay2018grad}  & \cmark & \xmark  & \cmark & \xmark &  \xmark \\
Relevance-CAM~\cite{lee2021relevance}  & \cmark & \xmark  & \cmark & \xmark &  \xmark \\
Integrated Grad-CAM~\cite{sattarzadeh2021integrated}  & \cmark & \xmark  & \cmark & \xmark &  \xmark \\

Rozenberg \etal~\cite{rozenberg2020localization}& \cmark & \cmark  & \xmark & \xmark &  \xmark \\
Karargyris \etal~\cite{karargyris2021creationeyegaze} & \cmark & \cmark  & \cmark & \xmark & \xmark \\
\midrule
\textbf{Ours} & \cmark & \cmark  & \cmark  & \cmark& \cmark \\
\bottomrule
\end{tabular}}
\vspace{-0.9em}
\label{tab:compare_sota_cap}
\end{table*}
Computer-aided diagnosis (CAD) has proven to be an invaluable tool in the medical field. In chest X-ray (CXR) diagnosis, the extensive growth of deep learning has given rise to several automated models that can outperform trained radiologists~\cite{rajpurkar2017chexnet, morid2021scoping}. In contrast to fully automatic systems, a good CAD framework should also improve the radiologist's performance~\cite{doi2007computer}. Despite steady improvements in deep learning methods in medical analysis \cite{bui2023sam3d, vo2023neural, tsuneki2022deep, egger2022medical, tran20223dconvcaps, le2021narrow, nguyen20213d, ho2021point, zhou2021deep}, there remains a problem: If the model makes a correct prediction, but the radiologist does not, then how does the system help the radiologist discern the truth? 
Sight is the essential first step in human thought, and radiologists must look carefully to verify whether there is an abnormality only after having extracted enough visual information~\cite{busby2018bias}. Thus, to assist radiologists effectively, we must address two crucial questions: where the radiologist should look, and how focused, or intense, they should. Answering these questions allows us to explore what findings can be diagnosed based on the radiologist's intensity.

The standard approach to address the first question is to visualize the internal features of the model using standard visualization methods, such as Class Activation Mapping (CAM) \cite{zhou2016learning,selvaraju2017grad,wang2020score}. However, many state-of-the-art techniques are heavily reliant on the usual black-box approach. The resulting heatmaps lack reliability as there is no constraint regarding any ground truth from physicians except the final disease label. Consequently, these approaches may make use of incorrect information, such as using the diaphragm as an indirect cue for cardiomegaly~\cite{karargyris2021creationeyegaze}. 
Other approaches simultaneously predict the disease and point out the location of it by making predictions in the form of bounding boxes ~\cite{rozenberg2020localization, liu2019align}. However, these approaches only address the first question of where the radiologist should look. To overcome these limitations, Karargyris \etal~\cite{karargyris2021creationeyegaze} introduces eye gaze datasets and modifies UNet~\cite{ronneberger2015unet} to generate heatmaps and predict abnormal findings. However, due to the bottleneck location prediction and classification both share, this approach encounters a significant challenge of relying on incorrect information when classifying. Clearly, to synchronously answer multiple questions  would require multiple heads in the model or multiple models creating different types of heatmaps.


To address all the above problems comprehensively, we propose a novel \textbf{unified controllable interpretable I-AI pipeline} for simultaneously generating radiologist-based anatomic attention heatmaps and predicting abnormal findings as illustrated in Figure \ref{fig:overview}. Our I-AI model takes a CXR image and anatomical prompt as inputs. To be controllable, our model first employs a short prompt specifying an anatomical part to guide the model's attention. To be interpretable, our model allows users to observe meaningful attention heatmaps of radiologists' explicit focus. I-AI model not only addresses the first question of localization but also captures the radiologists' focus intensity. Once obtaining where and how intense the radiologist gazes, our I-AI eliminates all extraneous information before predicting any abnormal findings, and therefore we ensure that our model cannot exploit erroneous data (i.e. diagnosis). This makes our network more interpretable and controllable compared to traditional black-box approaches. The I-AI model capacity comparison between our proposed method and related approaches is given in Table \ref{tab:compare_sota_cap}.

\vspace{-0.2em}

To obtain radiologist's intensity, we utilize the REFLACX dataset~\cite{lanfredi2021reflacx}, which contains a plethora of eye gaze information captured by high sensitivity hardware of radiologists analyzing CXR images. 
However, aligning a gaze sequence with an abnormal finding is non-trivial. For example, the radiologist's gaze can shift from the heart to the upper left lung, then to the right lung, and go back to the heart, and then they start to say "the heart size is normal". The randomness in the provided gaze sequence makes it hard to manually decide exactly which gaze points contribute to the diagnosis. To handle this, we propose a semi-automatic approach to extract gaze information using anatomical parts of the lung. Using this filtered data, we train, test our method and further verify a hypothesis that the classifier can achieve strong performance even without using the full image.

Our contribution can be summarized as follows:
\begin{itemize}[noitemsep,topsep=0pt]
    \item We propose I-AI, a novel unified controllable \& interpretable approach that uses a CXR image in conjunction with an anatomical prompt to determine the location and intensity of a radiologist's focus followed by the prediction of a corresponding finding. To the best of our knowledge, our method is the first in the medical domain to learn from radiologist-based anatomical gaze heatmap while offering controllability. 
    \item To train our I-AI model, we present a semi-automatic approach to extract radiologist-based anatomic heatmap from eye gaze datasets by using transcript and anatomic segmentation masks.
    \item We have conducted an extensive experiments and comparison to demonstrate the effectiveness of the proposed I-AI.
\end{itemize}

\begin{figure*}[t]
    \centering
    \includegraphics[width=0.9\linewidth]{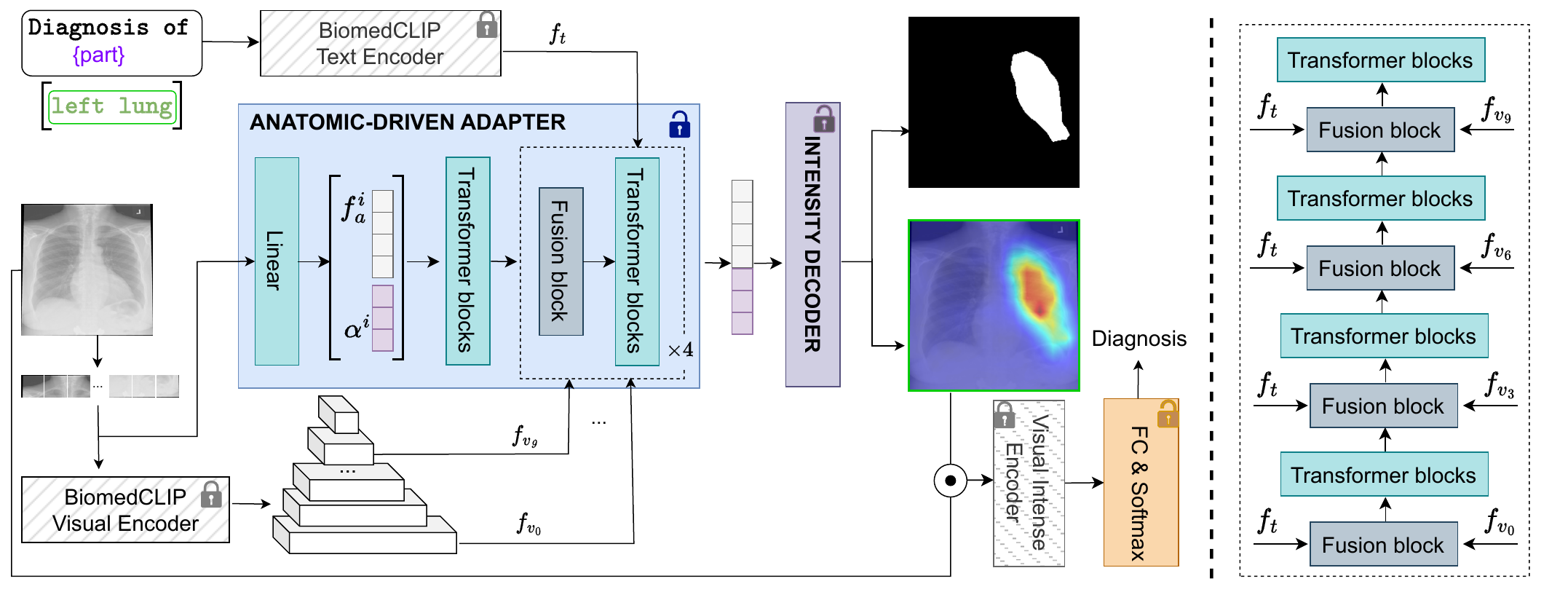}
    \caption{The detailed pipeline of our proposed controllable \& interpretable framework to decode radiologist's intense focus for accurate CXR diagnoses. In our framework, the frozen pre-trained model still serves to extract text embedding and CXR visual encoding, and the anatomic-driven adapter generates mask and attention heatmap to guide the deeper layers of the pre-trained model.}
    \vspace{-0.5em}
    \label{fig:tsan.detail} 
\end{figure*}

\section{Related works}
\textbf{Explainable Deep Learning.} Understanding a model's decision-making process holds significant importance today, particularly in CAD. Recent developments such as Class Activation Mapping (CAM)~\cite{selvaraju2017grad, sattarzadeh2021integrated, chattopadhay2018grad, lee2021relevance} have showcased one common approach: training a black box model and subsequently employing CAM-related techniques to visualize critical areas. While black-box models often exhibit high performance, they are recognized for their unreliability, as highlighted in literature~\cite{rudin2019stop}. Therefore, we design our model as an interpretable approach.


\textbf{Interpretable Deep Learning.}
Unlike the aforementioned explainable tools, a more desirable approach entails the design of a system wherein decisions are intrinsically linked to explainability, particularly in high-stakes medical contexts~\cite{rudin2019stop}. Generally, interpretable models aim to transform inputs into human-interpretable representations such as concepts or prototypes, which are then harnessed for prediction. To imbue the model with self-explanatory capabilities, many researchers have embraced the successful prototype-based approach~\cite{chen2019looks, nauta2021neural, nauta2023pip,ukai2022looks,rymarczyk2022interpretable, wang2021interpretable,rymarczyk2021protopshare}.
For instance, ProtoPNet~\cite{chen2019looks} introduces a prototypical part network that identifies prototypical parts within input images, leveraging this insight for the final prediction. PIP-net~\cite{nauta2023pip} learns prototypes that align closely with human visual perception, serving as scoring sheets during classification. In a supervised conceptual framework, TCAV~\cite{kim2018interpretability} is trained on data representing specific concepts. Adhering to the principles of interpretable models, we extend this methodology by enlisting the expertise to annotate the raw gaze sequences of a radiologist with three intentions, corresponding to three anatomical regions: left lung diagnosis, right lung diagnosis, and heart diagnosis. Subsequently, the model employs this intention-based information to diagnose the presence of anomalies.


\textbf{Disease condition localization.} Some works~\cite{liu2019align, han2022knowledge,  rozenberg2020localization} predict a bounding box to localize diseases with the ground truth being a bounding box and disease label. There are other works~\cite{zhu2022multi, ouyang2020learning} that train the model primarily on an image-level label and extract saliency maps or use Class Activation Mapping (CAM) to obtain the location of the disease. Karargyris \etal~\cite{karargyris2021creationeyegaze} predict a heatmap, but the ground truth is a full gaze map, and their GradCAM visualization indicates that the model is unreliable as it incorporates unrelated information in classification and heatmap prediction. Unlike previous works, we utilize a unique combination of eye tracking information, the reading of the radiologist, and anatomic segmentation to generate anatomic radiologist-based heatmaps.

\textbf{CXR disease classification.} Disease classification using CXR images has gained much attention recently. The earliest of these efforts, ChexNet~\cite{rajpurkar2017chexnet}, is a DenseNet~\cite{huang2017densenet} that uses the CXR image as its direct input. Since then, many efforts that use deep learning have risen from the related areas of supervised learning~\cite{yao2018weakly, taslimi2022swinchex}, semi-supervised learning~\cite{liu2020semi, liu2021self}, and self-supervised learning~\cite{gazda2021self, azizi2021big}. Besides using the full image to predict a disease, numerous studies~\cite{liu2019align, yan2018weakly, li2018thoracic, van2023probabilistic} 
suggest that location information of the disease can help in classification tasks. To the best of our knowledge, no existing methods use anatomic radiologist-based heatmaps in aiding and masking out irrelevant pixels in the image for classification. 




\section{Methodology: Proposed I-AI}
\label{sec:explainmodel}
\subsection{Problem Formulation}

Given a CXR image $x$ and an anatomical prompt $p$, i.e. "\texttt{Diagnosis of \{\}}" as a prefix with "\texttt{left lung}", "\texttt{right lung}", or "\texttt{the heart}", our goal is to produce a radiologist-based attention heatmap $a$ and corresponding label $y$.

Generally speaking, a radiologist-based attention heatmap should match the location and intensity of radiologists' eye gaze patterns and highlight relevant areas of the chest X-ray (CXR) for accurate diagnoses, as described in \cref{sec:data}. It should also derive the predicted label from a comparable amount of visual information that a radiologist would consider.

\subsection{Architecture}
\label{sec:arch}
To capture the textual modality while maintaining a good mask prediction, we design our heatmap predictor to be a lightweight \textit{Anatomic-Driven Adapter} that leverages the BiomedCLIP~\cite{zhang2023biomedclip} checkpoint, which was trained on 15 million image-caption pairs in PMC-15M~\cite{zhang2023biomedclip}, followed by a \textit{classifier}. The architecture is described in Figure \ref{fig:tsan.detail}.

\noindent
\textbf{Visual encoding.} First, the image will be split into $16 \times 16$ patches. Then, we feed the patches into the BiomedCLIP Visual Encoder. Following Xu \etal~\cite{xu2023san}, we extract the intermediate features $f_{v_i} \in \mathbb{R}^{H/k \times W/k \times 768}$ from 4 layers, i.e., \texttt{stem}, $3$, $6$, and $9$. 

\noindent
\textbf{Text encoding.} Unlike visual encoding, the anatomical prompts are short and concise, so the latent features of intermediate transformer layers are not meaningful for us. Therefore, we get only the final embedding $f_t \in \mathbb{R}^{512}$ from the BiomedCLIP Text Encoder module.

\noindent
\textbf{Anatomic-Driven Adapter.} Inspired by Xu \etal~\cite{xu2023san}, we train a vision transformer (ViT)~\cite{dosovitskiy2020vit} as an Anatomic-Driven Adapter by using the domain feature from BiomedCLIP from different scales. First, the input image $x$ is split into multiple $16 \times 16$ patches. Then, we use a linear projection to produce $f_a \in \mathbb{R}^{(14 * 14) \times D}$, where $D$ is the hidden dimension. We then concatenate $f_a$ with a scaling vector $\alpha \in \mathbb{R}^{ D}$. 
Next, we feed the concatenated feature into multiple stacked combinations of transformer layers and fusion blocks. Specifically, we fuse the feature from layers $\{\texttt{stem}, 1, 2, 3\}$ of our adapter ViT with $\{\texttt{stem}, 3, 6, 9\}$ layers in BiomedCLIP, a 12-layer ViT-B/16, i.e. \texttt{stem} to \texttt{stem}, $3$ to $1$, $6$ to $2$, and $9$ to $3$. For each fusion step, we also feed the text embedding $f_t$ into the fusion block as illustrated in Figure \ref{fig:tsan.detail} (right).

The intuition of including the scaling vector $\alpha$ is that each element in the last latent feature does not contribute equally across all anatomic parts, so the learnable scaling vector $\alpha$ allows the model to flexibly re-weight the last feature in the most suitable way to produce the final intense heatmap. 

\begin{figure}[t]
\begin{subfigure}{0.49\linewidth}
    \centering
    \includegraphics[width=\linewidth]{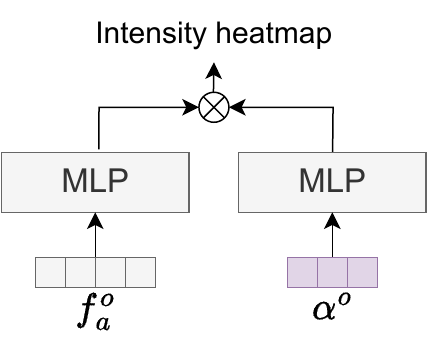}
    \caption{}
    \label{fig:heatmap-decoder}
\end{subfigure}
\unskip\ \vrule\
\begin{subfigure}{0.49\linewidth}
    \centering
    \includegraphics[width=\linewidth]{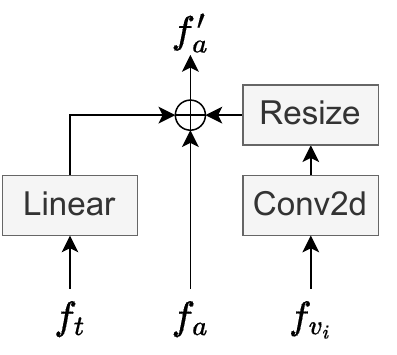}
    \caption{}
    \label{fig:fusion-block}
\end{subfigure}
\caption{The detailed illustration of our proposed modules: (a) Intensity Decoder module takes the output [$f^o_a$ $\alpha^o$] from Anatomic-Driven Adapter as its input and results in intensity heapmap; (b) Fusion block consists of three inputs of text embedding $f_t$, visual encoding $f_{v_i}$ and adapter latent feature $f_a$. $\bigoplus$ denotes element-wise addition.}
\end{figure}

\noindent
\textbf{Fusion block.} The fusion block has 3 inputs, the BiomedCLIP visual encoding at $i^{th}$ block $f_{v_i} \in \mathbb{R}^{H/k \times W/k \times 768}$, the BiomedCLIP text embedding $f_t \in \mathbb{R}^{768}$, and the adapter latent feature $f_a \in \mathbb{R}^{(14 * 14) \times D}$. For the visual encoding $f_{v_i}$, we first use a convolution layer to reduce the channel dimension, and then we perform an interpolation operation to resize the resolution to create $f_{v_i}' \in \mathbb{R}^{(14 * 14) \times D}$. On the other side, we pass $f_t$ through a linear layer to project it into the fusion space with dimension of $D$ to create $f_t' \in \mathbb{R}^{D}$. After that, we add them together
\begin{equation}
    f_a' = f_a + f_{v_i}' + f_t'
\end{equation}
where the add operation of $f_t'$ is broadcasting. A fusion block is shown in Figure \ref{fig:fusion-block}.

We use the add operation for feature fusion: a simple and strong established baseline~\cite{xu2023san}. While other fusion mechanisms may enhance performance, they are beyond the scope of this paper.

\noindent
\textbf{Intensity Decoder.} The Intensity Decoder receives the output of the last layer of our adapter to generate the heatmap, i.e. latent feature $f_a^o \in \mathbb{R}^{(14 * 14) \times D}$ and scaling vector $\alpha^o \in \mathbb{R}^{D}$. We first pass those two features into two separated multilayer perceptrons (MLPs). We then use matrix multiplication between them to produce a small gray-scale attention logit $a_l \in \mathbb{R}^{ (14 * 14)}$. To get the final attention logit, we resize $a_l$ into $\hat{a_l} \in \mathbb{R}^{ W \times H}$.  In our implementation, we set $D$ to $240$. Fig.~\ref{fig:heatmap-decoder}  illustrates Intensity Decoder module.

\noindent
\textbf{Heatmap loss.} Given the predicted logit $\hat{a_l} \in \mathbb{R}^{W \times H}$ and ground truth heatmap $a \in \mathbb{R}^{W \times H}$, we compute the $L_2$ loss:
\begin{equation}
    L_2 = \|a_l - \sigma^{-1}(a)\|_2
\end{equation}
where $\sigma^{-1}(x) = \ln{\frac{x}{1-x}}$ is the logit function. Note that we compute the loss before applying the sigmoid function to the predicted logit heatmap to avoid the issue of vanishing gradients.

\noindent
\textbf{Mask-related losses.} Given the predicted logit $\hat{a_l}$ and ground truth heatmap $a$, we use binary cross entropy loss and dice loss on the masks created from $a_l$ and $a$. First, we apply the sigmoid function $\sigma(x)=1/(1+\exp(-x))$ on the predicted logit to create the predicted heatmap $\hat{a}$. We then define a function $f(\cdot)$: 
\begin{equation}
    f(a_{i,j}) = \begin{cases} 1 & \text{ if } a_{i,j} > 0 \\ 0 &\text{ otherwise} \end{cases}
\end{equation}
We apply $f(\cdot)$ on all values of $a$ and $\hat{a}$ to create the ground truth mask $m$ and predicted mask $\hat{m}$. Finally, we apply the standard dice loss $L_{dice}$ and binary cross entropy loss $L_{ce}$ as in~\cite{cheng2022masked}.
The the final loss for training the Anatomic-Driven Adapter is
\begin{equation}
    L_h = \lambda_1 L_2  + \lambda_2 L_{ce} + \lambda_3 L_{dice}
    \label{eq:final_loss}
\end{equation}
where the weights $\lambda_1, \lambda_2, \lambda_3$ are all set to $1.0$.

\noindent
\textbf{Classifier.}
Using the predicted heatmap $\hat{a} \in [0,1]^{W \times H}$ from the previous step, we multiply $\hat{a}$ with the image $x$ element-wise to re-weight the importance of all pixels. Afterwards, we use a Visual Intense Encoder and Linear layer followed by Softmax activation as Classifier to extract and predict the finding label $y$. In our implementation, the Visual Intense Encoder is the BiomedCLIP Visual Encoder. Finally, we use a cross entropy loss to guide the classifier.

\section{Data preparation}
\label{sec:data}

\subsection{Settings}
REFLACX~\cite{lanfredi2021reflacx} provides eye gaze data for more than 2,500 CXRs from MIMIC-CXR~\cite{johnson2019mimic}, where each gaze sequence is captured using a device with sensitivity of $1000$Hz.  
However, REFLACX does not provide a gaze map for each anatomic part of the lung. To construct the disease-level gaze heatmap ground truth, we manually annotate the data. The process of creating the ground truth is discussed in~\cref{sec:gt_making,sec:class}. Note that the only category that has more than $300$ samples after annotating is Cardiomegaly. 
Therefore, Cardiomegaly is treated as a separate subset, while all other diseases are categorized into left or right lung subsets.
After labeling the data, we split it into four distinct settings below 
\begin{itemize}[noitemsep,topsep=0pt]
    \item C: Only samples with verbal transcript that specifically mentions cardiomegaly.
    \item L: Only samples with transcript that specifically mentions left lung. 
    \item R: Only samples with transcript that specifically mentions right lung. 
    \item M: Merging all samples from C, L, and R.
\end{itemize}

For each subset, we split 70\% for training, 15\% for evaluation, and 15\% for testing. We also keep the balance between positive and negative ratio to be 1:1. The data distribution is shown in Table \ref{tab:data_analysis}. 

\begin{table}[t]
\centering
\caption{Data distribution corresponding to four distinct settings: C: cardiomegaly, L: Left lung, R: Right lung, M: entire chest and merging all samples from the C, L, and R subsets.}
\begin{tabular}{l|c}
\toprule
\textbf{Settings} & \textbf{No. samples} (train:val:test)  \\
\midrule
C      & 611:131:132             \\
L      & 631:143:145     \\
R      & 575:129:125     \\
M      & 1817:403:402  \\ 
\bottomrule
\end{tabular}
\label{tab:data_analysis}
\end{table}

\subsection{Ground truth heatmap} 
\label{sec:gt_making}
\begin{figure}[t]
    \centering
    \includegraphics[width=\linewidth]{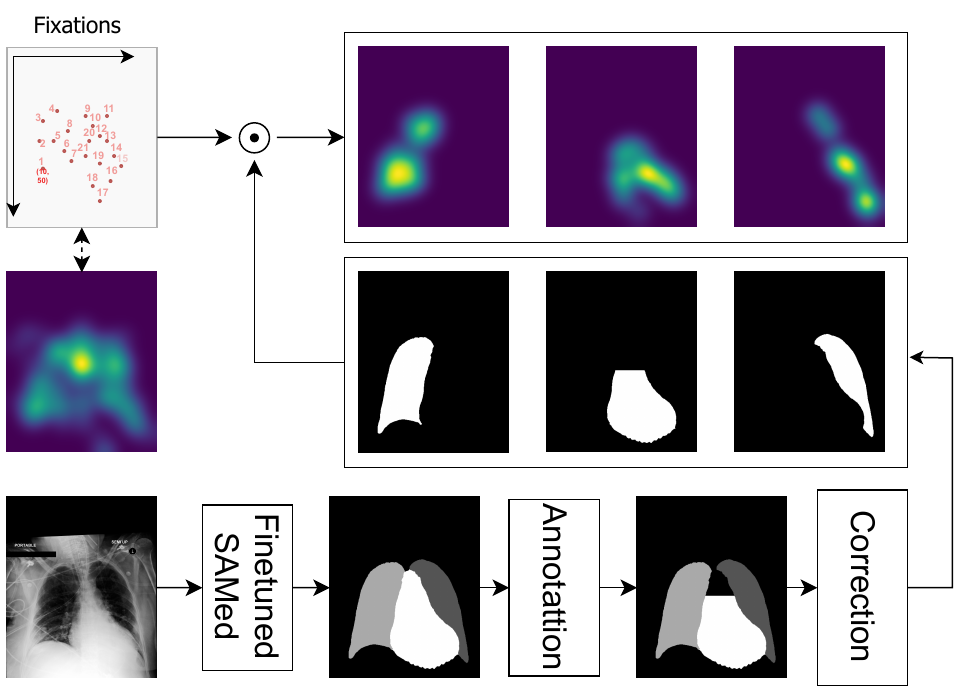}
    \caption{The pipeline of creating ground truth gaze map from eye gaze dataset.}
    \label{fig:new_heatmap_gt}
\end{figure}

To create the ground truth heatmap, we perform two steps: make anatomic masks and filter fixations. Figure \ref{fig:new_heatmap_gt} demonstrates the overall pipeline for making ground truth gaze heatmaps.

\noindent
\textbf{Anatomic masks.} REFLACX also does not provide anatomic masks, so we have to create these masks as well. Currently, the anatomic masks for three big parts are provided by EGD-CXR~\cite{karargyris2021creationeyegaze}: left lung, right lung, and the mediastinum. We finetune SAMed~\cite{zhang2023samed} on EGD-CXR, then use the finetuned model to make inferences on REFLACX. Then, we manually correct the segmentation masks if there is any problem. For example, we heuristically cut out the top one third of each mediastinum mask to make the heart masks, but automatic script may cut too much, so we have to fix it.

\noindent
\textbf{Filtering fixation sequence.} For a particular anatomic region, we can acquire the fixations by looking for keywords in the provided transcripts. For example, \texttt{cardiomegaly} or \texttt{enlarged} and \texttt{cardiac} for setting C.
Then, we will pick the rightmost sentence to decide the upper end of the interval containing our desired fixations.
Specially, given a sequence of sentences $\{s_1, s_2, \dots, s_{n}\}$, if we find $s_3, s_4$ and $s_{10}$ contain the keyword, we will use $s_{10}$. As a result, the chosen fixations are in interval $[0, e]$, where $e$ is the ending time of $s_{10}$. Using the predicted mask from before, we exclude any fixation point located beyond its boundaries. Note that the starting time of $0$ is required to capture potentially relevant visual information from the moment the radiologist takes their very first glance.  
Finally, by applying a Gaussian filter with radius of $150$ on the chosen fixations' coordinates, we obtain the final ground truth heatmap. 
More details can be found in our Supplementary Material.



\noindent
\textbf{Anatomical prompt.} We also need the input prompt to guide the model. For our anatomical prompt, we use the prefix "\texttt{diagnosis of \{\}}".
After the prefix, we append our target: "\texttt{left lung}" for left lung heatmap prediction, "\texttt{right lung}" for right lung heatmap, and "\texttt{heart}" for heart heatmap. 

In order to ensure the validity of the results and the effectiveness of the automated process, all corrections are meticulously examined and carried out by expert radiologists.

\subsection{Classification} 
\label{sec:class}
Based on our four distinct settings, we design four yes/no questions for classifying findings:
\begin{itemize}[noitemsep,topsep=0pt]
    \item C: Is there cardiomegaly?
    \item L: Is there a finding (excluding Cardiomegaly) in the left lung of the image?
    \item R: Is there a finding (excluding Cardiomegaly) in the right lung of the image?
    \item M: Is there a finding in the masked image?
\end{itemize}

\begin{figure*}[t] 
   \centering
  \huge
\resizebox{\linewidth}{!}{
\setlength{\tabcolsep}{2pt}
\begin{tabular}{cccccccccc}
\rotatebox[origin=l]{90}{\hspace{-0.5cm} \textbf{Relevance-CAM~\cite{lee2021relevance}}} &
\shortstack{\includegraphics[width=0.33\linewidth]{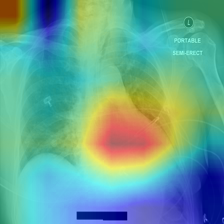}}&
\shortstack{\includegraphics[width=0.33\linewidth]{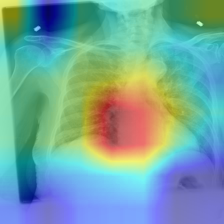}}&
\shortstack{\includegraphics[width=0.33\linewidth]{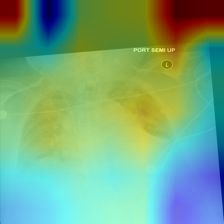}}&
\shortstack{\includegraphics[width=0.33\linewidth]{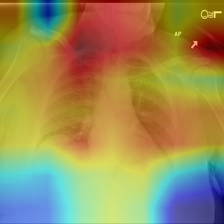}}&
\shortstack{\includegraphics[width=0.33\linewidth]{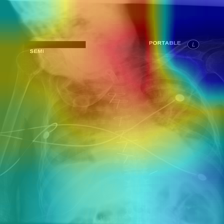}}&
\shortstack{\includegraphics[width=0.33\linewidth]{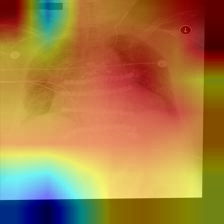}}&
\shortstack{\includegraphics[width=0.33\linewidth]{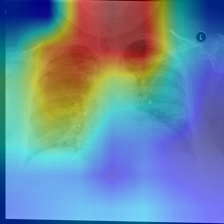}}&
\shortstack{\includegraphics[width=0.33\linewidth]{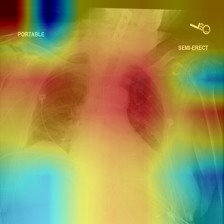}}&
\shortstack{\includegraphics[width=0.33\linewidth]{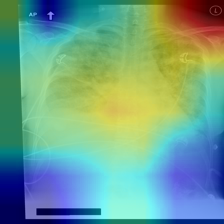}}\\[1pt]
\rotatebox[origin=l]{90}{\hspace{0.1cm}  \textbf{Grad-CAM~\cite{selvaraju2017grad}}} &
\shortstack{\includegraphics[width=0.33\linewidth]{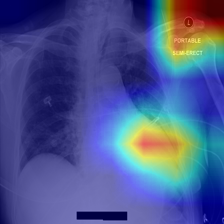}}&
\shortstack{\includegraphics[width=0.33\linewidth]{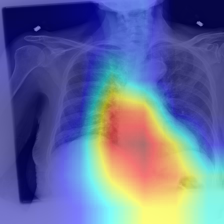}}&
\shortstack{\includegraphics[width=0.33\linewidth]{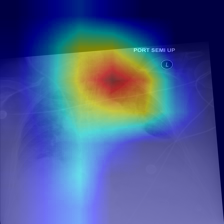}}&
\shortstack{\includegraphics[width=0.33\linewidth]{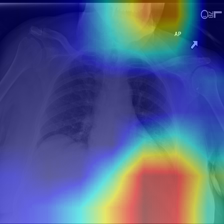}}&
\shortstack{\includegraphics[width=0.33\linewidth]{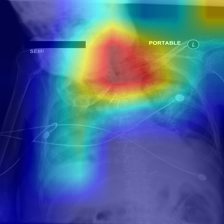}}&
\shortstack{\includegraphics[width=0.33\linewidth]{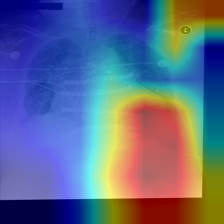}}&
\shortstack{\includegraphics[width=0.33\linewidth]{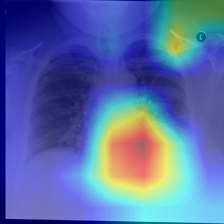}}&
\shortstack{\includegraphics[width=0.33\linewidth]{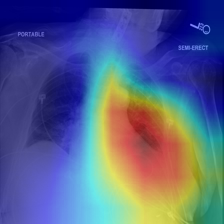}}&
\shortstack{\includegraphics[width=0.33\linewidth]{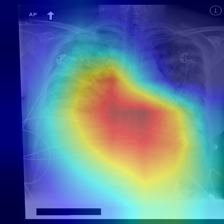}}\\[1pt]
\rotatebox[origin=l]{90}{\hspace{0.1cm}  \textbf{Grad-CAM++~\cite{chattopadhay2018grad}}} &
\shortstack{\includegraphics[width=0.33\linewidth]{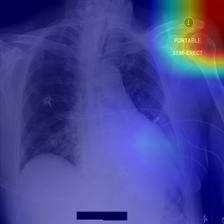}}&
\shortstack{\includegraphics[width=0.33\linewidth]{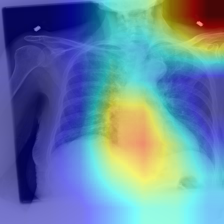}}&
\shortstack{\includegraphics[width=0.33\linewidth]{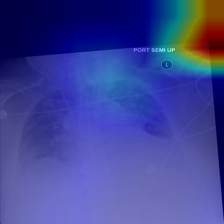}}&
\shortstack{\includegraphics[width=0.33\linewidth]{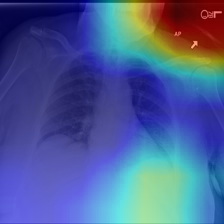}}&
\shortstack{\includegraphics[width=0.33\linewidth]{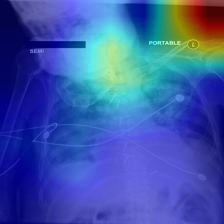}}&
\shortstack{\includegraphics[width=0.33\linewidth]{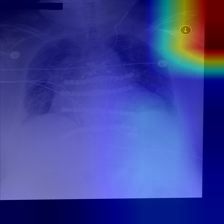}}&
\shortstack{\includegraphics[width=0.33\linewidth]{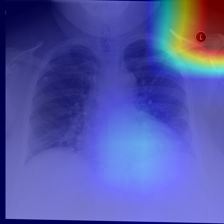}}&
\shortstack{\includegraphics[width=0.33\linewidth]{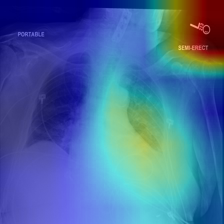}}&
\shortstack{\includegraphics[width=0.33\linewidth]{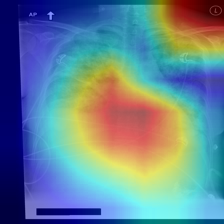}}\\[1pt]

\rotatebox[origin=l]{90}{\parbox{5.6cm}{\centering  \textbf{Integrated}\\ \textbf{Grad-CAM~\cite{sattarzadeh2021integrated}}}} &
\shortstack{\includegraphics[width=0.33\linewidth]{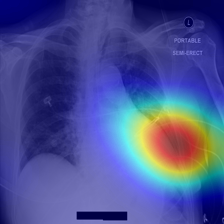}}&
\shortstack{\includegraphics[width=0.33\linewidth]{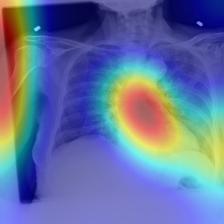}}&
\shortstack{\includegraphics[width=0.33\linewidth]{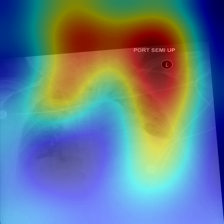}}&
\shortstack{\includegraphics[width=0.33\linewidth]{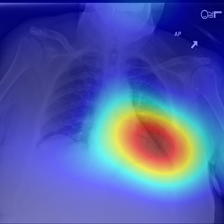}}&
\shortstack{\includegraphics[width=0.33\linewidth]{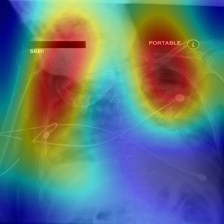}}&
\shortstack{\includegraphics[width=0.33\linewidth]{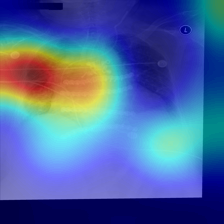}}&
\shortstack{\includegraphics[width=0.33\linewidth]{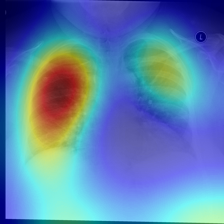}}&
\shortstack{\includegraphics[width=0.33\linewidth]{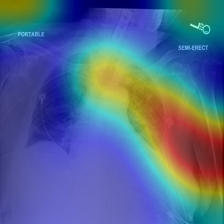}}&
\shortstack{\includegraphics[width=0.33\linewidth]{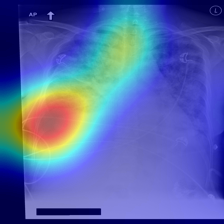}}\\[1pt]

\rotatebox[origin=l]{90}{\parbox{5.6cm}{\centering  \textbf{Grad-CAM \\ \hspace{0.0cm}\ Karargyris \etal  \\\cite{karargyris2021creationeyegaze}}}} &
\shortstack{\includegraphics[width=0.33\linewidth]{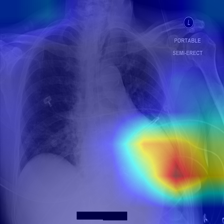}}&
\shortstack{\includegraphics[width=0.33\linewidth]{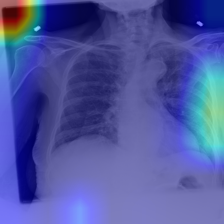}}&
\shortstack{\includegraphics[width=0.33\linewidth]{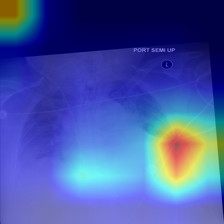}}&
\shortstack{\includegraphics[width=0.33\linewidth]{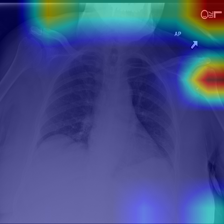}}&
\shortstack{\includegraphics[width=0.33\linewidth]{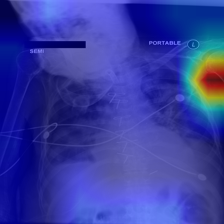}}&
\shortstack{\includegraphics[width=0.33\linewidth]{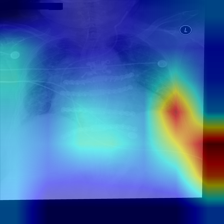}}&
\shortstack{\includegraphics[width=0.33\linewidth]{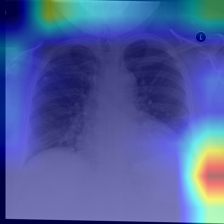}}&
\shortstack{\includegraphics[width=0.33\linewidth]{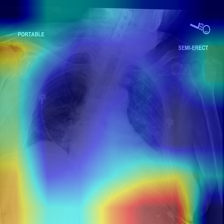}}&
\shortstack{\includegraphics[width=0.33\linewidth]{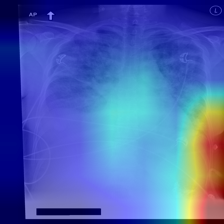}}\\[1pt]

\rotatebox[origin=l]{90}{\parbox{5.6cm}{\centering\hspace{0.0cm}  \textbf{Karargyris \etal \\\cite{karargyris2021creationeyegaze}}}} &
\shortstack{\includegraphics[width=0.33\linewidth]{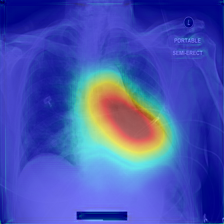}}&
\shortstack{\includegraphics[width=0.33\linewidth]{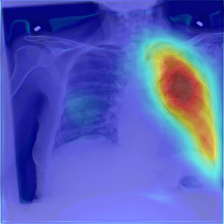}}&
\shortstack{\includegraphics[width=0.33\linewidth]{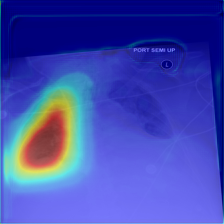}}&
\shortstack{\includegraphics[width=0.33\linewidth]{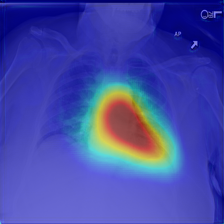}}&
\shortstack{\includegraphics[width=0.33\linewidth]{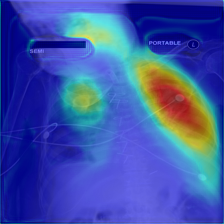}}&
\shortstack{\includegraphics[width=0.33\linewidth]{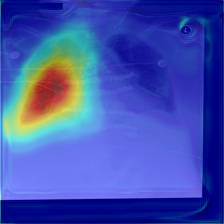}}&
\shortstack{\includegraphics[width=0.33\linewidth]{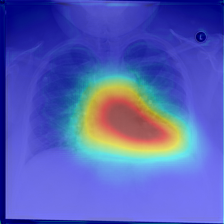}}&
\shortstack{\includegraphics[width=0.33\linewidth]{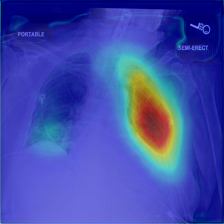}}&
\shortstack{\includegraphics[width=0.33\linewidth]{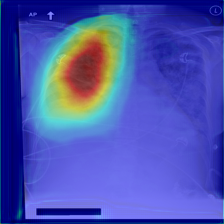}}\\[15pt]
\hline
\\
\rotatebox[origin=l]{90}{\hspace{0.2cm} \textbf{Ground truth}} &
\shortstack{\includegraphics[width=0.33\linewidth]{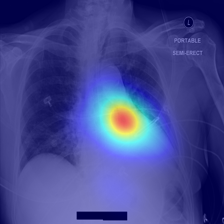}}&
\shortstack{\includegraphics[width=0.33\linewidth]{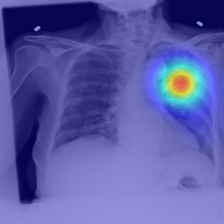}}&
\shortstack{\includegraphics[width=0.33\linewidth]{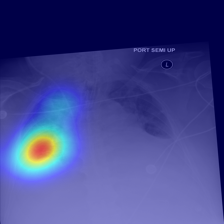}}&
\shortstack{\includegraphics[width=0.33\linewidth]{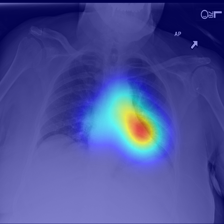}}&
\shortstack{\includegraphics[width=0.33\linewidth]{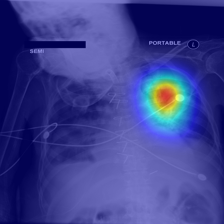}}&
\shortstack{\includegraphics[width=0.33\linewidth]{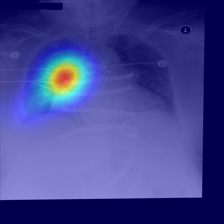}}&
\shortstack{\includegraphics[width=0.33\linewidth]{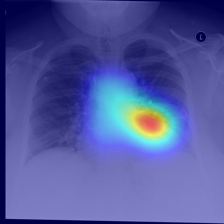}}&
\shortstack{\includegraphics[width=0.33\linewidth]{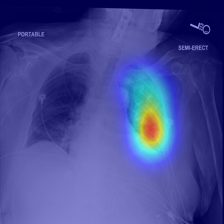}}&
\shortstack{\includegraphics[width=0.33\linewidth]{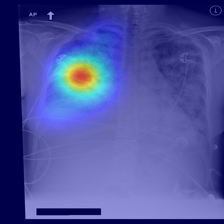}}\\[15pt]
\hline
\\
\rotatebox[origin=l]{90}{\hspace{1.7cm} \textbf{I-AI (Ours)}} &
\shortstack{\includegraphics[width=0.33\linewidth]{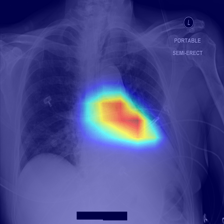}}&
\shortstack{\includegraphics[width=0.33\linewidth]{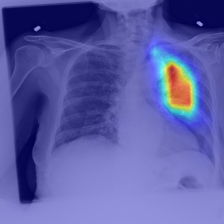}}&
\shortstack{\includegraphics[width=0.33\linewidth]{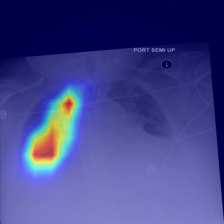}}&
\shortstack{\includegraphics[width=0.33\linewidth]{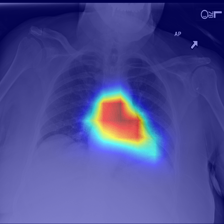}}&
\shortstack{\includegraphics[width=0.33\linewidth]{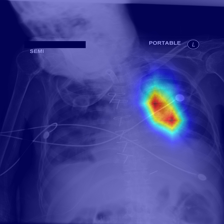}}&
\shortstack{\includegraphics[width=0.33\linewidth]{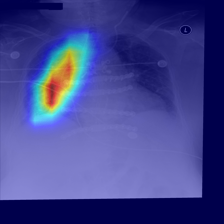}}&
\shortstack{\includegraphics[width=0.33\linewidth]{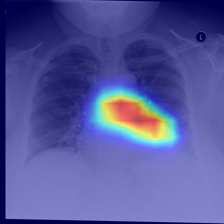}}&
\shortstack{\includegraphics[width=0.33\linewidth]{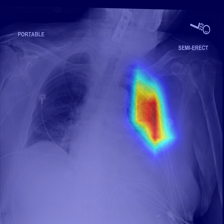}}&
\shortstack{\includegraphics[width=0.33\linewidth]{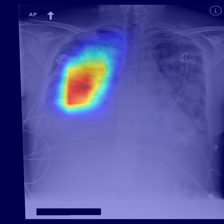}}\\[1pt]

\end{tabular}
}
    \caption{Comparison of the results from various CAM methods, Karargyris \etal~\cite{karargyris2021creationeyegaze}, and ours. Relevance-CAM~\cite{lee2021relevance}, Grad-CAM~\cite{selvaraju2017grad}, and Grad-CAM++~\cite{chattopadhay2018grad} disappoint with unreliable heatmaps. Integrated Grad-CAM~\cite{sattarzadeh2021integrated} exhibits some promise despite imprecision and error. Karargyris \etal~\cite{karargyris2021creationeyegaze} performs well. However, our approach stands out as the closest match to the ground truth.}
    \label{fig:2DCompare}
\end{figure*}

\section{Experiments and Results}
\subsection{Experiment settings}
\noindent
\textbf{Implementation details.} The ViT adapter is an $8$-layer vision transformer with dimension of $240$, $6$ attention heads, and an input patch size of $16 \times 16$. The BiomedCLIP's visual encoder is a $12$-layer ViT-B/16 pretrained on resolution $224^2$. The BiomedCLIP's text encoder is a $12$-layer BERT with a vocab size of $30,522$. We freeze both the text encoder and visual encoder of BiomedCLIP in the heatmap prediction and classification stages. The MLPs in the Intensity Decoder have $3$ fully connected layers and hidden dimension of $256$. We proceed to train them with a learning rate of $0.0001$, batch size of $16$, $60,000$ iterations, and AdamW optimizer~\cite{loshchilov2017adamw}. The training process takes roughly 4 hours on a single Quadro RTX 8000 GPU. The hyper-parameter and fusion layer choices are described in Supplementary Materials.

\noindent
\textbf{Comparison.} For the black-box approaches, we train a ResNet-101~\cite{he2016resnet} on our settings. Then we use Relevance-CAM~\cite{lee2021relevance}, Grad-CAM~\cite{selvaraju2017grad}, Grad-CAM++~\cite{chattopadhay2018grad}, and Integrated Grad-CAM~\cite{sattarzadeh2021integrated} to get the performance of various CAM methods. For the heatmap prediction approach, we use Karargyris \etal~\cite{karargyris2021creationeyegaze} and TransUNet~\cite{chen2021transunet} to compare with our proposed method. Note that, both ResNet-101 and Karargyris \etal are trained on separated settings, i.e. C, L, and R subsets, because we have one input with three outputs. Then, we take the average of all subsets to get the final scores. Meanwhile, our proposed method is trained on only M subset.

\noindent
\textbf{Metrics.} To quantify the performance at capturing radiologist's intensity, we use the mean of $L_2$ (mL2), $L_1$ (mL1), Structural SIMilarity (mSSIM), and peak signal-to-noise ratio (mPSNR) over all samples. On the other hand, we also need to measure how well the heatmap can filter out irrelevant pixels by using intersection over union on foreground (fgIoU) and background (bgIoU). Additionally, we also use Frequency Weighted IoU (fwIoU). 

\subsection{Qualitative results}

Figure \ref{fig:2DCompare} shows the difference of our results compared to other results from Karargyris \etal~\cite{karargyris2021creationeyegaze} and CAM methods. Despite being trained on 3 different subsets, the CAM methods produce bad and unreliable heatmaps because we do not constrain them. Note that, although Karargyris \etal's predicted heatmap is not too far off and its accuracy is 75\% (Table \ref{tab:all_classifier}), its Grad-CAM visualizations show that Karargyris \etal is using mostly irrelevant information to classify and produce heatmaps. Unlike the aforementioned results, our method produces more precise heatmaps thanks to the radiologist-based heatmap constraint. 

Moreover, we can also see from Figure \ref{fig:2DCompare} that Karargyris \etal~\cite{karargyris2021creationeyegaze} results are bigger than ours. Therefore, it has a better chance at covering the ground truth heatmap and has a better chance at achieving a high fgIoU score with a high false positive rate. However, the score for the intensity should not be as high because there are many regions that should be paid much less attention. 


\subsection{Quantitative Results}
\label{sec:class_result}
\cref{tab:heatmap_result} shows that our method achieves superior performance over other heatmap generators. Among the CAM-based methods, Integrated Grad-CAM is the highest scoring, but its scores are still lower than methods that directly predict heatmaps. For instance, Integrated Grad-CAM has a fgIoU score lower than ours by approximately $25$ units. In terms of "where to look at", Karargyris \etal's IoU scores closely match our method. In particular, ours has a slightly lower fgIoU than Karargyris \etal by $1.3$, but our method has a better bgIoU by approximately $9$ unit. 

In regards to intensity-type metrics, our method is superior in all metrics. Specifically, our methods outperform Karargyris \etal with $+0.28$ mSSIM, $+4.49$ mPSNR, $-0.08$ m$L_1$, and $-0.03$ $mL_2$.
This agrees with our visual analysis in \cref{fig:2DCompare}.

\begin{table*}[t]
    \centering
    \caption{Performance comparison with state-of-the-art methods. Our I-AI stands out for its fine-grained localization and precision. Note that Grad-CAM (Karargyris \etal~\cite{karargyris2021creationeyegaze}) method extracts the heatmap from Karargyris \etal~\cite{karargyris2021creationeyegaze} using Grad-CAM.}

    \resizebox{0.7\linewidth}{!}{%
    \begin{tabular*}{0.8\linewidth}{@{\extracolsep{\fill}}lccccccc}
    \toprule
       
    \multirow{2}{*}{\textbf{Methods}} & \multicolumn{3}{c}{\textbf{Location}} & \multicolumn{4}{c}{\textbf{Intensity}} 
    \\ 
    \cmidrule(lr){2-4} \cmidrule(lr){5-8} 
                                   & 
                                   fgIoU$\uparrow$ & 
                                   bgIoU$\uparrow$ &
                                   fwIoU$\uparrow$ & mSSIM$\uparrow$ & mPSNR$\uparrow$ & mL1$\downarrow$ & 
                                   mL2$\downarrow$ 
                                   \\
    \midrule
        Relevance-CAM~\cite{lee2021relevance}  & 15.49 & 40.00 & 37.25  &  0.10 & 5.64 & 0.50 & 0.29  \\ 
        Grad-CAM~\cite{selvaraju2017grad}  & 18.91 & 78.12 & 71.49 & 0.24 & 10.40 & 0.22 & 0.11 \\ 
        Grad-CAM++~\cite{chattopadhay2018grad}  & 8.76 & 79.85 & 71.88  &  0.20 & 10.69 & 0.20 & 0.09  \\ 
        Integrated Grad-CAM ~\cite{sattarzadeh2021integrated}  & 12.27 & 82.44 & 74.58  &  0.37 & 12.48 & 0.17 & 0.07 \\ 
        Grad-CAM (Karargyris \etal~\cite{karargyris2021creationeyegaze})  & 6.68 & 54.81 & 49.97 & 0.36 & 9.62 & 0.27 & 0.12 \\ 
        Karargyris \etal~\cite{karargyris2021creationeyegaze}  & \textbf{39.59} & 83.69 & 79.26  &  0.55 & 13.77 & 0.16 & 0.05 \\ 
        TransUNet~\cite{chen2021transunet} & 33.68 & 90.13 & 84.54 & 0.83 & 12.79 & 0.09 & 0.06 \\  	
        \midrule
        \textbf{I-AI (Ours)} & 37.27 & \textbf{92.44} & \textbf{86.96} & \textbf{0.83} & \textbf{18.26} & \textbf{0.08} & \textbf{0.02} \\
    \bottomrule
    \end{tabular*}}
    \vspace{-0.57em}
    \label{tab:heatmap_result}
\end{table*}

\cref{tab:all_classifier} shows that our pipeline achieves the highest accuracy at 76\%, despite using only a portion of the input image based on our predicted attention heatmap. Note that our accuracy is similar to Karargyris et al.'s, implying that using a radiologist-based heatmap to mask the input image does not harm, and might even enhance overall performance.

\begin{table}[t]
\centering
\caption{Accuracy comparison between all classifiers.}
\begin{tabular}{lc}
\toprule
\textbf{Model}   & \textbf{Accuracy}(\%) \\
\midrule
Resnet-101       & 71.64            \\
Karargyris \etal       & 75.12  \\
TransUNet & 74.88 \\ 
\midrule
I-AI (Ours)       & 76.86\\
\bottomrule
\end{tabular}
\vspace{-0.3em}
\label{tab:all_classifier}
\end{table}

\vspace{-0.3em}

\subsection{Ablation study}

\noindent
\textbf{Heatmap prediction of particular setting.} \cref{tab:heatmap_ablation1} shows the robustness of our model across all settings. Our model achieves high performance with marginal difference between the full setting (M) versus subsetting C, L, and R. 

\begin{table}[t]
\centering
\caption{Ablation study: Heatmap prediction of particular settings.}
\begin{tabular*}{\linewidth}{@{\extracolsep{\fill}}lcccc}
\toprule
\textbf{Settings}   & fwIoU$\uparrow$ & mSSIM$\uparrow$ & mPSNR$\uparrow$ & mL1$\downarrow$\\
\midrule
C     & 87.42 & 0.84 & 18.80 & 0.07             \\
L      & 86.51 & 0.85 & 18.76 & 0.08          \\
R      & 87.16 & 0.82 & 17.48 & 0.10         \\
M   & 86.96 & 0.83 & 18.26 & 0.08 \\ 
\bottomrule
\end{tabular*}
\vspace{-0.3em}
\label{tab:heatmap_ablation1}
\end{table}



\noindent
\textbf{The importance of mask-related losses.} During training process, we notice that the gradient flow of $L_2$ is not enough for the model to learn where to look, and it can easily collapse to a local minima where a metric like mL2 is good, but other metrics like fgIoU are bad. As shown in \cref{tab:heatmap_ablation3}, fgIoU dramatically drops to $4.06$, while mL2 is $0.03$. Therefore, we use masks created from the heatmaps together with cross entropy loss and dice loss to guide the model.

\begin{table}[t]
    \centering
    \caption{Ablation study: the impact of losses on heatmap predictor.}
    \resizebox{0.97\linewidth}{!}{%
    \begin{tabular*}{\linewidth}{@{\extracolsep{\fill}}lcccccc}
    \toprule
        \multicolumn{3}{c}{\textbf{Losses}} & \multicolumn{2}{c}{\textbf{Location}} & 
        \multicolumn{2}{c}{\textbf{Intensity}} \\ 
        \cmidrule(lr){1-3} \cmidrule(lr){4-5} \cmidrule(lr){6-7}
        
        $L_2$ &  $L_{ce}$ & $L_{dice}$ & 
        {fgIoU$\uparrow$} & {fwIoU$\uparrow$}  & {mPSNR$\uparrow$} & {mL2$\downarrow$} \\ 
        \midrule
          \cmark &  \xmark &  \xmark &  4.06 &  81.80   &  16.62 &  0.03 \\ 
        \xmark& \cmark & \cmark &  38.16 & 85.04   & 13.11 &  0.04 \\ 
        \cmark & \cmark &\cmark & 37.27 & 86.96 & 18.26  & 0.02 \\ 
        \bottomrule
    \end{tabular*}
    }
    \vspace{-0.3em}
\label{tab:heatmap_ablation3}
\end{table}


\noindent
\textbf{The importance of scaling vector $\alpha$.} We define a learnable scaling vector $\alpha$ in \cref{sec:arch} to help the model learn. From the output $f_a^o \in \mathbb{R}^{W \times H \times D}$, it is true that we can naively create the final output by taking the mean of the last dimension. However, as shown in \cref{tab:heatmap_ablation4}, the inflexibility of naively averaging the feature space effectively prevents the model from learning. 

\begin{table}[t]
    \centering
    \caption{Ablation study: the scaling vector on anatomic-driven heatmap predictor.}
    \begin{tabular*}{\linewidth}{@{\extracolsep{\fill}}llllllll}
    \toprule
        \textbf{Settings} & fwIoU$\uparrow$ & mSSIM$\uparrow$ & mPSNR$\uparrow$ & mL1$\downarrow$ \\ 
        \midrule
w/o $\alpha$     & 62.87 & 0.31 & 12.59  & 0.17           \\
w/ $\alpha$      & 86.96 & 0.83 & 18.26 & 0.08          \\
        \bottomrule
    \end{tabular*}
    \vspace{-0.55em}
\label{tab:heatmap_ablation4}
\end{table}

\noindent
\textbf{The importance of radiologist-based heatmap in classification.}
As shown in \cref{tab:classifier_ablation}, the classifier can be improved by using the ground truth heatmap. The area ratio is defined as $\frac{H}{T}$, where $H$ is the number of heatmap values larger than $0$, and $T$ is the number of pixels. Even though the predicted heatmaps cover more area, the performance does not scale accordingly. We can see that correctly identifying where and how long to look is more beneficial than simply covering a larger part of the image.

\begin{table}[t]
\centering
\caption{Ablation study: classification performance using predicted gaze map versus ground truth gaze map.}
\resizebox{0.8\linewidth}{!}{%
\begin{tabular}{llcc}
\toprule
\textbf{Settings} &  \textbf{Heatmap}  & \textbf{Area ratio} (\%) & \textbf{Accuracy}(\%) \\
\midrule
\multirow{2}{*}{C} & Ground truth &  21.49  & 83.33            \\
                   & Ours   & 43.31 & 80.30             \\  \midrule 
\multirow{2}{*}{L} & Ground truth & 16.43  & 81.37              \\
                   & Ours   & 43.14  & 74.48              \\ \midrule
\multirow{2}{*}{R} & Ground truth & 18.18  & 80.00              \\
                   & Ours    & 43.95  & 75.20               \\ \midrule
\multirow{2}{*}{M} & Ground truth & 18.18 & 81.34              \\
                   & Ours    & 43.49  & 76.86             \\ 
\bottomrule
\end{tabular}}
\vspace{-0.5em}
\label{tab:classifier_ablation}
\end{table}


\section{Conclusion}
We present I-AI, a novel unified controllable \& interpretable pipeline to decode and reconstruct radiologists' intense focus and diagnosis from CXR. 
Our I-AI model can simultaneously address three critical questions: where a radiologist looks, how long a radiologist focuses on specific areas, and what findings a radiologist diagnoses.
Our I-AI achieves effective interpretability by aligning the output (findings) with intermediate layers (heatmap) and controllability through prompt-guided intensity generation and finding classification.
Extensive experiment shows the superiority of our I-AI approach compared to other methods, even when utilizing only a portion of the image. This highlights the importance of focusing on the most relevant regions rather than processing the entire input indiscriminately.

\section*{Acknowledgment}
This material is based upon work supported by the National Science Foundation (NSF) under Award No OIA-1946391 RII Track-1, NSF 1920920 RII Track 2 FEC, NSF 2223793 EFRI BRAID, NSF 2119691 AI SUSTEIN, NSF 2236302.

\clearpage
{\small
\bibliographystyle{ieee_fullname}
\bibliography{egbib}
}

\end{document}